\documentclass[runningheads]{llncs}
\usepackage{graphicx}
\usepackage{booktabs}
\usepackage{multirow}
\usepackage{subfig}
\usepackage{hyperref}
\usepackage[final]{trackchanges}

\begin{document}
%
\title{Implementation of a perception system for autonomous vehicles using a~detection-segmentation network in SoC FPGA}
\titlerunning{Perception system using a det-seg network in SoC FPGA}
%

\author{Maciej Baczmanski\inst{1} \and
Mateusz Wasala\inst{1} \href{https://orcid.org/0000-0002-8631-8428}{\includegraphics[width=16pt]{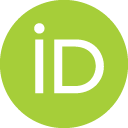}} \and
Tomasz Kryjak\inst{1}  \href{https://orcid.org/0000-0001-6798-4444}{\includegraphics[width=16pt]{images/orcid.png}}
}

\authorrunning{M. Baczmanski et al.}

\institute{Embedded Vision Systems Group, Computer Vision Laboratory, \\ Department of Automatic Control and Robotics, \\ AGH University of Krakow, Al. Mickiewicza 30, 30-059 Krakow, Poland \\
\email{mbaczmanski@student.agh.edu.pl \\
\{mateusz.wasala, tomasz.kryjak\}@agh.edu.pl}
}
\maketitle              
\begin{abstract}



Perception and control systems for autonomous vehicles are an active area of scientific and industrial research.
These solutions should be characterised by \remove{both} high efficiency in recognising obstacles and other environmental elements in different road conditions, real-time capability, and energy efficiency.
Achieving such functionality requires an appropriate algorithm and a suitable computing platform.
In this paper, we have used the MultiTaskV3 detection-segmentation network as the basis for a perception system that can perform both functionalities within a single architecture.
It was appropriately trained, quantised, and implemented on the AMD Xilinx Kria KV260 Vision AI embedded platform.
By using this device, it was possible to parallelise and accelerate the computations. 
Furthermore, the whole system consumes relatively little power compared to a CPU-based implementation (an average of 5 watts, compared to the minimum of 55 watts for weaker CPUs, and the small size (119mm x 140mm x 36mm) of the platform allows it to be used in devices where the amount of space available is limited. It also achieves an accuracy higher than $97\%$ of \add{the} mAP (mean average precision) for object detection and above $90\%$ of \add{the} mIoU (mean intersection over union) \remove{score} for image segmentation.
The article also details the design of the Mecanum wheel vehicle, which was used to test the proposed solution in a mock-up city.

\end{abstract}
\section{Introduction}
\label{sec:intro}

Today, we are witnessing the rapid development of advanced mobile robotics, including autonomous cars and drones (unmanned aerial vehicles, UAV).
This would not be possible without advances in the implementation of perception and control systems, including the use of deep neural networks (DNN).
DNNs make it possible to achieve high accuracy, but memory and computational complexity remain significant challenges.
In order to meet the requirements of mobile platforms, i.e. low latency and low energy consumption, it becomes necessary to use specialised hardware platforms such as SoC FPGAs (System on Chip Field Programmable Gate Arrays) or eGPUs (embedded Graphic Processing Units).
These solutions also have the advantage of relatively small size and weight.
It is also worth noting that a major challenge is the reliability analysis of network-based solutions, including their explainability \cite{xai}.
This is of particular importance when traffic safety, for example, depends on DNNs detections or control.


In perception systems, two basic tasks can be roughly distinguished: object detection and segmentation (semantic and instance).
Object detection is the marking of objects belonging to the considered classes (e.g. cars, pedestrians, cyclists, traffic signs, etc.) in the image with bounding boxes or sometimes binary masks.
Semantic segmentation involves assigning to each pixel a label that tells what object it belongs to (e.g. drivable area, horizontal road sign, vegetation, buildings, persistent, or sky).
Instance segmentation, on the other hand, allows different labels to be given to pixels belonging to two separate objects of the same class (e.g. two pedestrians).
It should be noted that object detection is a simpler and thus computationally less complex task. 
A typical solution /change{using}{that uses} DNNs is the YOLO (You Only Look Once) family of algorithms \cite{yolo_survey}.
In contrast, segmentation, especially of instances to obtain similar information, is much more complex -- requiring both longer learning and inference.
U-Nets \cite{unet} are typically used for semantic segmentation and Mask R-CNN-based \cite{maskrcnn} solutions for instances.


For autonomous vehicle perception systems, the tasks of detection and segmentation appear together.
For objects such as pedestrians, vehicles, bicycles, vertical road signs, or traffic lights, the use of detection is sufficient.
However, for the detection of drivable area or horizontal road signs (including pedestrian crossings), it is better to use segmentation.
Hence, detection-segmentation networks have been proposed in the literature, which combine the advantages of both approaches and, at the same time, thanks to a common backbone (encoder), are characterised by lower computational complexity and an easier learning process than instance segmentation approaches.
A detection-segmentation network, in addition to the aforementioned backbone, consists of a segmentation head and several detection heads.
Examples of such networks are YOLOP \cite{YOLOP}, HybridNets \cite{HybridNet} and MultiTask V3 \cite{mtv3} discussed in Section \ref{sec:prvious_work}.


Taking into account the properties of the detection segmentation networks discussed above, we decided to use this solution as the basis for the perception system of our autonomous vehicle model.
We used the MultiTask V3 network, which we implemented and deployed on two embedded platforms: SoC FPGA Kria KV260 and an eGPU (NVIDIA Jetson Nano and Xavier NX).
The experiments performed showed that detection-segmentation networks represent a good compromise between accuracy, performance, and power consumption.
We also discussed the design of the Mecanum wheeled vehicle used.
To the best of our knowledge, this is the first paper that discusses the hardware implementation of a perception system based on a detection-segmentation network implemented in an SoC FPGA, the results of which were applied to the control of an autonomous vehicle model.


The remainder of this paper is structured as follows. 
In Section \ref{sec:prvious_work} we discuss the relevant prior works on detection-segmentation networks and DNNs acceleration on SoC FPGA.
Section \ref{sec:methods} discusses the methods used, including the hardware implementation of the considered DNNs, and the design of the autonomous vehicle model.
The results obtained are summarised in Section \ref{sec:experiments}. 
The paper ends with conclusions and a discussion of possible future research.

\section{Previous work}
\label{sec:prvious_work}


Three types of deep neural networks can be distinguished in current vision systems: detection, segmentation, and detection-segmentation.
As mentioned in the introduction, detection-segmentation networks represent a compromise between the accuracy of instance segmentation and the speed of simple detection and are therefore an interesting solution for autonomous vehicle perception systems.
Several architectures of detection-segmentation networks have been proposed in the literature.


The first is YOLOP \cite{YOLOP}.
It allows object detection and segmentation of drivable area and horizontal road markings.
It consists of a common encoder and 3 separate decoders (one for detection and two for segmentation).
It has been trained and evaluated on the popular \textit{BDD100k} dataset \cite{bdd100k}.
The second is HybridNets \cite{HybridNet}, which is very similar to YOLOP in terms of functionality.
It consists of 4 components: encoder (EfficientNet V2 architecture), neck, detection head (inspired by YOLOv4), and segmentation head.
The \textit{BDD100k} dataset was also used for training and evaluation.
The third architecture, used in this work, is the MultiTask V3 \cite{mtv3} proposed by AMD Xilinx.
It is worth noting that it is included in the Vitis AI library as a demonstrator of its capabilities, but to our knowledge, it has not been described in a scientific publication.
Details of its construction are presented in Section \ref{subsec:seg_det_fpga}.
Unlike YOLOP and HybridNets, it also includes a depth estimation module.
However, it has not been evaluated on a publicly available dataset.


The topic of hardware acceleration of deep neural networks, especially for embedded computing, is the subject of intense academic and industrial research due to its very high practical importance.
A whole spectrum of solutions is encountered, from dedicated chips for AI acceleration (e.g. Intel Neural Compute Stick, Google Coral, Tesla FSD Chip), through programmable SoC FPGAs to eGPU platforms.
A detailed overview of the solutions is beyond the scope of this article, and we refer interested readers, for example, to the review \cite{embeddedAI} or the work \cite{nn_fpga_survey}.


In this work, we have chosen to use an SoC FPGA platform and also run the selected network on an eGPU platform for comparison.
Reprogrammable devices have been a proven platform for implementing vision algorithms for years, which was the main reason for our choice.
In addition, they tend to have lower power consumption than eGPUs.
Of the available detection-segmentation networks, we chose MultiTask V3 for two reasons.
First, from our previous experiments, it had the highest efficiency and relatively low computational complexity for our scenario.
Second, it was well-prepared by AMD Xilinx for acceleration in SoC FPGAs, which facilitated its use in the target perception and control system.

\section{Implementation of the perception and control system}
\label{sec:methods}


\begin{figure}[!t]
\centering
\begin{tabular}{cc}
\subfloat[]{\includegraphics[width=0.45\textwidth,height=4.5cm]{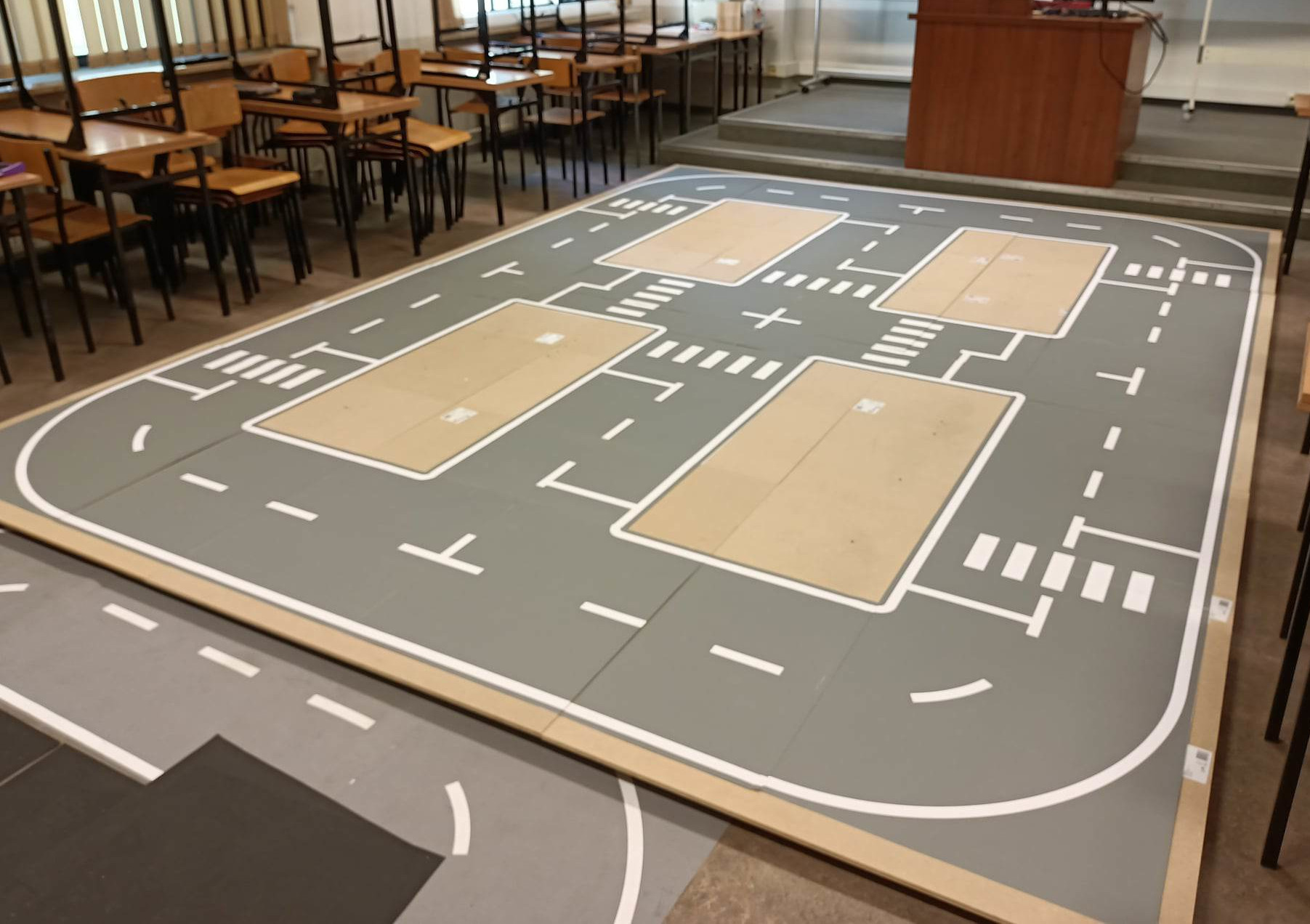}\label{fig:makieta}} &
\subfloat[]{\includegraphics[width=0.45\textwidth,height=4.5cm]{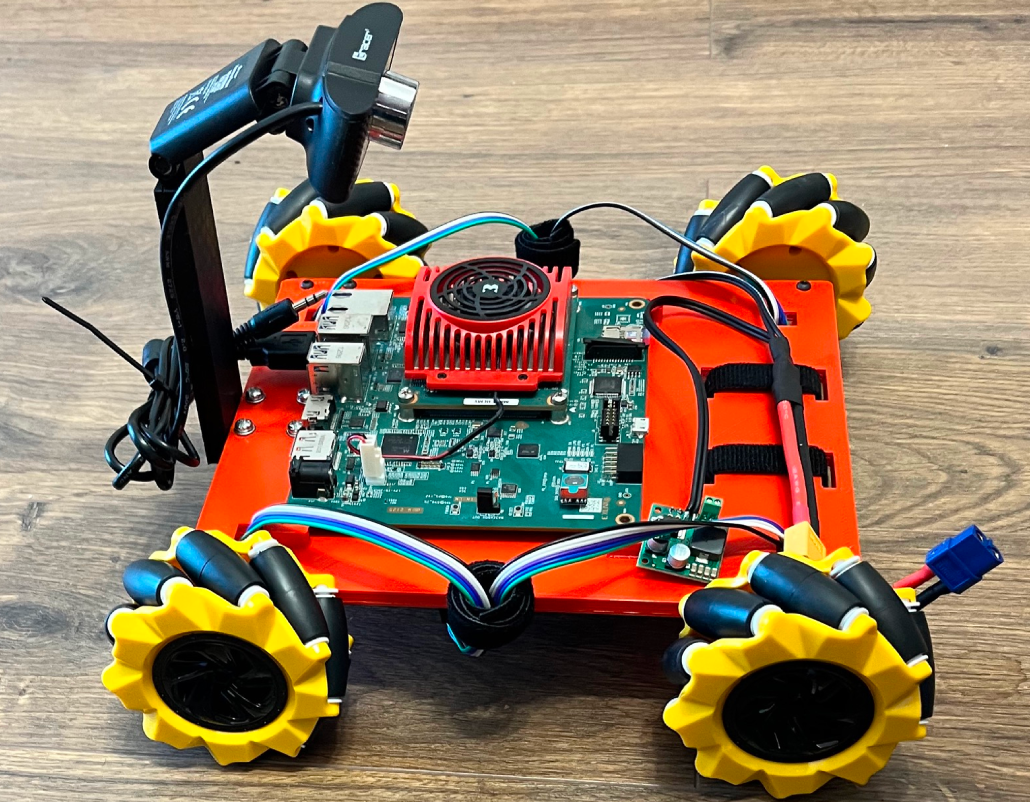}\label{fig:pojazd}}
\end{tabular}

\caption{The mock-up of a city made by us \protect\subref{fig:makieta} and the model of an autonomous vehicle  \protect\subref{fig:pojazd} with Mecanum wheels and all equipment.}
\label{fig:mockup_car_model}
\end{figure}

The starting point for our research was the FPT'22 \cite{FPT} competition, the aim of which is to create a model of an autonomous vehicle capable of driving according to the road traffic rules in a mock-up city.
Figure \ref{fig:makieta} shows the used mock-up city.
It is equipped with horizontal markings (traffic lanes, pedestrian crossings), traffic lights, figures imitating pedestrians, and various objects (obstacles) to be avoided on the road.
Thanks to this test environment, it is possible to evaluate the perception and control system of an autonomous vehicle.
The research presented can be divided into four phases: the design and construction of an autonomous vehicle equipped with Mecanum wheels, the design of electronics and assembly equipment, the implementation of the perception and control algorithm on the AMD Xilinx Kria KV260 platform, and the programming of a low-level algorithm to control the motors for the Mecanum wheels.
The most important part of the work is the implementation of the perception and control system.
It uses a detection-segmentation deep convolutional neural network architecture that is parallelised, quantised, and accelerated on an embedded SoC FPGA platform.
On the other hand, the Mecanum wheels allow for precise manoeuvring, and the detection-segmentation network provides the necessary information about obstacles and other elements of the environment.
In addition, the PID controller implemented in the motor controllers ensures stable driving, which is essential for the safety of the vehicle.

\subsection{Detection-segmentation network in SoC FPGA}
\label{subsec:seg_det_fpga}

\begin{figure}[!t]
\centerline{\includegraphics[width=1\textwidth]{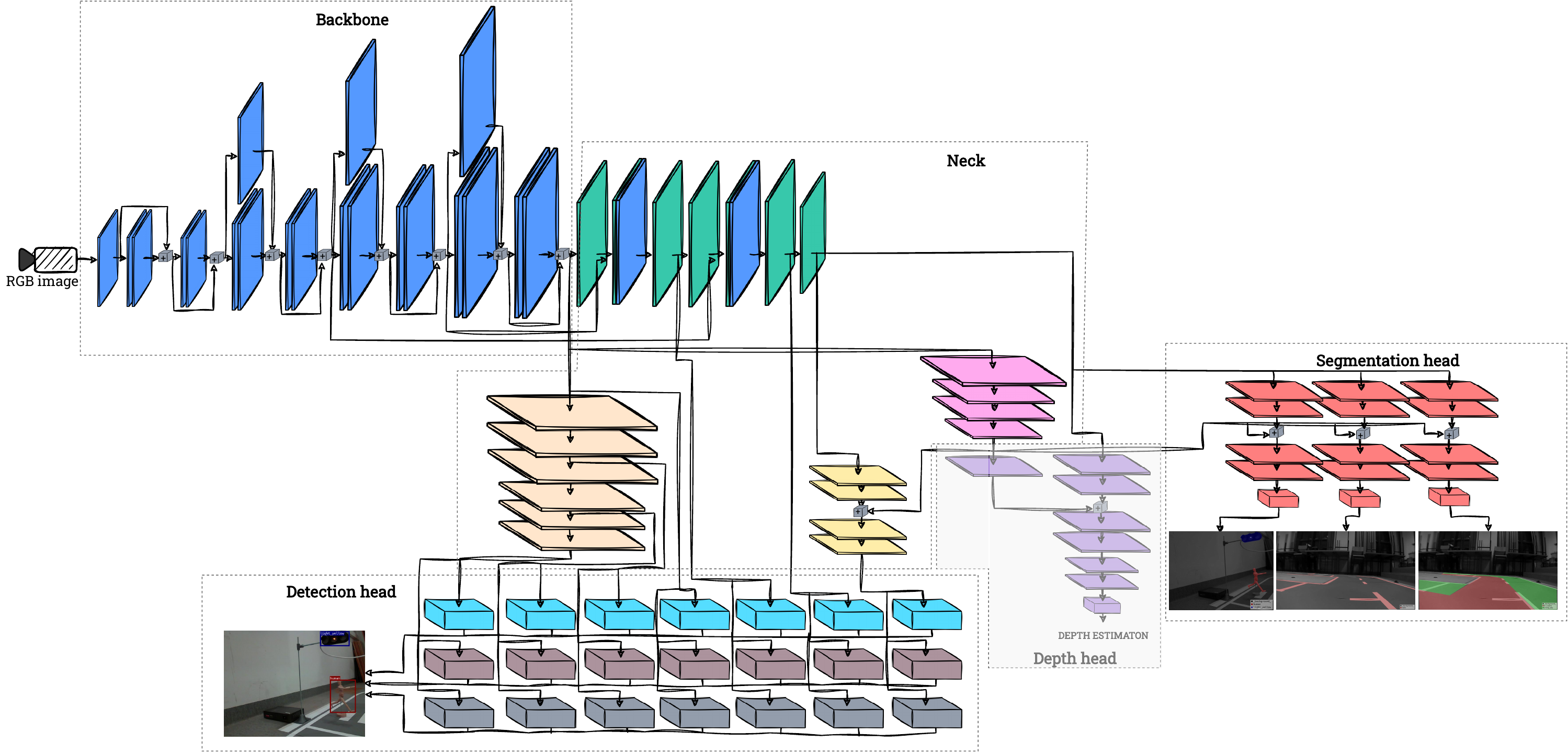}}
\caption{Scheme of the MultiTask V3 deep neural network, showing layers of neurons grouped into sections. An input image is processed within successive layers to extract features. The features are used to generate output data: detections, segmentation, and also a depth map.}
\label{fig:multitask_scheme}
\end{figure}

MultiTask V3 is a deep convolutional neural network, designed by the developers of Vitis AI (AMD Xilinx) as part of an open source library made available for the development process \cite{mtv3}\footnote{MultiTask V3 has not been described in a published scientific paper.}.
Its architecture is shown in Figure \ref{fig:multitask_scheme} and allows the simultaneous execution of five tasks: detection, three types of segmentation, and depth estimation (not used in this work).

The segmentation part of the architecture is divided into three branches. 
Each branch can focus on a different task, such as segmenting detected objects, lanes (drivable area), or road markings.
This approach makes it easier to prepare training sets, as these can be separated from each other, allowing a pixel to be classified in more than one class (e.g. a road marking should still be detected as a lane).
The additional use of detection means that an in-depth analysis of detected objects (e.g. in terms of shape or occupied area in the image) is optional and performed only in special cases.
The MultiTask V3 network architecture consists of several elements.
First, the input image is transferred to the \textit{Backbone} segment, which is used for feature extraction.
This is based on the ResNet-18 convolutional neural network.
Then, thanks to the use of encoders and convolutional layers, the \textit{Neck} segment allows further feature extraction and the combination of low-level and high-level features.
The features obtained are transferred to the appropriate branches: \textit{Detection}, \textit{Depth}, and \textit{Segmentation} heads.
In them, again, thanks to the use of convolution, activation operations, and normalisation, the corresponding result tensors are generated.

Due to the specificity of the project and the high complexity of the training set for depth estimation, the \textit{Depth} head training was not considered.
For the remaining branches, three training sets were prepared, one common for object detection and segmentation and two for drivable area segmentation and road markings.
The data for the training sets were obtained from recordings made on a city mock-up, which made it possible to prepare them strictly for the assumed task.
250 photos were obtained for the set containing the detected objects and 500 photos for the set showing the drivable area.
The images were then manually labelled using the LabelMe software.
\add{To speed up the process of labelling lanes, a preprocessing script was created, which used binarisation with a given threshold to create initial masks of lane markings, which then could be corrected manually in the software.
This method could be useful to speed up the process of labelling for bigger datasets for real-life deployment.}
The generated datasets were converted into a format compatible with the framework used to train the network.
The framework is open source, based on Python, uses the PyTorch libraries, and is published in the Vitis AI libraries.
As the software was written for older versions of the libraries and Python, corrections had to be made in order for the code to run properly.
Once the modifications had been made, the software was launched using the prepared datasets.
The model was trained using the GTX 1060 M GPU on sets split 80/20 between training and validation. 
The training was stopped after 450 epochs if there was no improvement in network performance.

The next step was to quantise the network model so that it could be run on an embedded SoC FPGA platform.
This was done using the software described above.
The quantisation is based on the vai\_p\_pytorch API provided by AMD Xilinx.
Finally, the model was compiled into an architecture-compatible format using the vai\_c\_xir program, also provided by AMD Xilinx.

The final detection-segmentation model has been launched on the Kria KV260 SoC FPGA platform \cite{kria_dim}.
Kria is designed for the development of advanced image processing applications, allowing the acceleration of neural networks thanks to the use of \change{DPU(Deep Processing Unit)}{a Deep Processing Unit (DPU)}.
The platform's operating system is Ubuntu, with PYNQ software installed, which allows a program to be created in Python on notebooks using the DPU overlay.
In addition, by using the WiFi USB adapter and modifying the operating system's network settings, it is possible to communicate with the platform via SSH (Secure Shell) and through the Jupyter Notebook server created, allowing the algorithm to be executed and its operation to be analysed in real-time.
This communication also makes it possible to continuously monitor the consumption of resources and the performance of the algorithm.
Thanks to the libraries used, it is possible to collect image frames from a connected USB camera with a resolution of $512 \times 320$ pixels, convert them into the network input tensor, and then analyse the output tensors using methods from the OpenCV library.
\add{DPU overlay's API accepts and returns only tensors of shape compatible with the model's input and output layers.
Preprocessing of input images, consisting of rescaling and changing colour models, and postprocessing, focused on gathering output data from raw tensors, were implemented in Python language.}
The implemented algorithm imports the necessary libraries and defines data pre-processing and processing functions.

\add{Additionally, a second version of the algorithm was implemented to run it on any Linux-based system with Pytorch library installed.
This version can be run on any device that supports USB ports, as it does not use any of Xilinx's provided libraries and relies on a trained model from before quantisation.
For evaluation purposes, the algorithm was run on NVIDIA Jetson (Nano and Xavier NX) platforms, which support computation parallelisation using eGPU.}

\subsection{Vehicle control algorithm}

The algorithm captures the last frame from the USB camera, pre-processes it (size, colour space), and converts it into tensors, which are then fed into the MultiTaskv3 neural network.
The network returns tensors which are then converted into masks: segmentation of detected objects, segmentation of drivable area, segmentation of road markings, and bounding boxes of detected objects.
The received data is then analysed: first, it is checked that the pedestrian or obstacle is not in the ROI\add{ (Region of Interest)}, which is defined as a short distance in front of the vehicle.
In the case of a pedestrian, the vehicle should stop, and in the case of an obstacle, the overtaking manoeuvre should be initiated.
The lines are then checked.
The detection of a continuous cross-line marking triggers a vehicle stop.
Based on the sideline, it is possible to determine the trajectory of movement.
If the sideline is not in the ROI -- on the left side of the image, the segmentation of the drivable area allows checking if the vehicle is at an intersection or in a curve, which means it needs to turn.
Based on the results of the analysis, a trajectory is determined and transmitted to the Arduino microcontroller, which controls the motors.
The loop then returns to the initial step and continues indefinitely.

\subsection{Hardware setup}
\label{subsec:hardware_setup}

The electronics project consisted of placing the Arduino Nano Every microcontroller, based on the ATMega4809, on the breadboard, allowing the use of hardware interrupts on any pin.
The microcontroller is directly connected to the motor encoders and four Pololu DRV8838 motor controllers, which allow control using the PWM (Pulse-width Modulation) signal.
The power section consists of a LiPo package and step-down converters: 12V for the FPGA platform and 6V for the motors.
The microcontroller communicates and is powered via a USB connection to the FPGA platform.
The motor control was programmed on the microcontroller in the language provided by Arduino, based on C++.
The program receives the set values from the FPGA platform through the UART protocol in the format $V_x, V_y,\omega $, where $V_x$ is the longitudinal velocity vector, $V_y$ is the transverse velocity vector, and $\omega$ is the given angular velocity of rotation relative to the geometric centre of the vehicle.
From the above values, the angular velocities set values for each of the motors are determined.
The rotation of each wheel changes the signals on the encoder connected to it.
Using hardware interrupts, it is possible to determine the angle that each of the motors has turned, which is counted in the counter assigned to it, and stored in the cache.
The interrupt timer has been implemented in the program, which calls the function exactly every 0.1 seconds.
This function retrieves the current counter reading and compares it with the previous one.
This is used to determine the angular velocity, the previous values of which are also stored and differentiated for the purposes of the PID (Proportional Integral Derivative) controller.
Then, for each motor, the control set for the given speed, the control error and its differential are determined, which makes it possible to determine the P and D terms of the PID controller.
The values obtained are used to determine the filling of the PWM signal sent to the motor controllers.
The program runs in an infinite loop, and in asynchronous mode, the microcontroller is constantly waiting for a new reference to be sent.

In order to better adapt the vehicle to the dimensions of the city mock-up, all its elements were made using 3D printing technology, such as adapters for the motors to mount the wheels, USB camera holder, base platform adapted to mount the motors, cameras, electronics, power supply, and the main computing platform.
Four Pololu HP micromotors with 150:1 gears and encoders were attached to the base platform, on which the shaft was mounted using Mecanum 80mm diameter wheels with adapters.
On the underside of the platform is a breadboard with electronics to control the motors and a 14.8V nominal LiPo pack.
At the top of the chassis \add{there} is a computer platform and a USB camera mount.
Figure \protect\ref{fig:pojazd} shows the model of the autonomous vehicle described above.

\section{Evaluation of the detection-segmentation network}
\label{sec:experiments}

The first experiment was to compare the quality (efficiency, accuracy) of network model inference before and after quantisation. 
The tests were performed using the libraries provided by AMD Xilinx, discussed earlier. 
Each branch was evaluated on the test set and the results are summarised in Tab\change{el}{le}s \ref{tab:detection}, \ref{tab:drivable_results}, \ref{tab:lane_results} and \ref{tab:seg_results}. 
As can be seen, quantisation resulted in a slight quality decrease (of the order of less than one per cent). 
This means, therefore, that the model used by the SoC FPGA platform will behave almost identically to the one run on a PC equipped with a graphics card in the environment provided by AMD Xilinx.
\add{The conducted experiment also shows that a prepared dataset of 750 pictures in total was enough to train the model for the given task.
It is worth noting that it was created for a small mock-up with repetitive scenery, so shooting more pictures could result in overfitting of the model.
The dataset seems to be a good base which could be extended for models that could be run on similar, but larger mock-ups.
For real-life deployment of the model, it would be worth considering using bigger datasets, which take into consideration environmental variability (different times of day, weather, etc.). 
There are some datasets available, such as BDD100K, or a custom dataset could be created, similarly as described in Section} \ref{subsec:seg_det_fpga}.


To test the efficiency and cost-effectiveness of the proposed solution, a series of performance tests were carried out on the Kria KV260 platform.
The input to the algorithm was a pre-prepared dataset derived from footage recorded on a mock-up of the city.
During operation, the use of the quad-core Cortex-A53 processor clocked at 1.3 GHz \remove{used in the platform}, the use of RAM (Random Access Memory) and CMA (Contiguous Memory Allocator), and the power consumption of the SOM (System on Module) platform were checked. 
The results are shown in Table \ref{tab:consumption}. 
It is worth noting that the platform makes full use of one CPU core. 
According to the manufacturer's documentation, it is possible to run the algorithm using multithreading, but this would involve higher power consumption.
The results show that the platform consumes only around 5W of power when running, which allows it to be considered energy efficient. 


In order to compare the performance of the platform used, the inference time of the MultiTask V3 network and the execution time of one iteration of the algorithm \change{was}{were} examined. 
The same algorithm was then run on the NVIDIA Jetson Nano and NVIDIA Jetson Xavier NX eGPU platforms, using the pre-quantisation model and the PyTorch library to run the network.  
The results of the algorithm's efficiency on the platforms are shown in Table \ref{tab:platforms}.


Experiments show that the Kria KV260 platform has demonstrated the best performance in its power consumption class. 
In terms of processing speed, it clearly outperforms the NVIDIA Jetson Nano platform, with the same power consumption.
It also runs faster than the NVIDIA Jetson Xavier NX platform in 10W consumption mode. 
Only when using the 20W consumption mode does the NX platform achieve approximately 0.5 fps (frames per second) more, but at the cost of four times higher power consumption.
\add{The results also show differences in the hardware implementation of the model on eGPU and DPU.
Compared to NVIDIA Jetson platforms with a power consumption of 10W or less, the inference time of the Kria KV260 is significantly faster.
Executing the neural network model on an FPGA is more efficient than using a GPU solution.}


The achieved processing speed of almost 5 FPS is sufficient for the algorithm to make a decision in a satisfactory time. 
However, the results show that the application of deep neural networks on energy-efficient embedded platforms is still a significant challenge.


To sum up. The best results were \change{obtained}{achieved} on the Kria KV260 SoC FPGA platform.
The SoC FPGA platform allows us to obtain satisfactory results in terms of accuracy, efficiency, and power consumption.
It should be noted that the currently implemented algorithm is still under development, and the results show that it would be beneficial to focus more on code optimisation and system reconfiguration to utilise all CPU cores. This could slightly increase power consumption, but even 10W of consumption can be considered low for a platform that would be the most important element of an autonomous car.
The code used in the experiments described is available at \url{https://github.com/vision-agh/mt_kria}.

\add{Implementation of the algorithm, using the Kria KV260 platform was tested on a constructed mock-up.
The experiment consisted of driving the car on the road, with figures of pedestrians and obstacles placed on it.
The robot correctly detected incoming events, such as approaching an intersection or road curve, pedestrians, and obstacles.
It reacted accordingly to the rules of the FPT'22 competition.
Results show that the implementation of an algorithm based on the segmentation-detection neural network can be used for autonomous robots and could be scaled for bigger autonomous vehicles operating on roads.}


\begin{table}[!t]
\setlength{\tabcolsep}{6pt}
\centering
\caption{Comparison of results for object detection (mAP -- mean Average Precision).}
\begin{tabular}{|c|c|c|c|}
\hline
Quantisation &  \multirow{2}{*}{mAP$_{50}$ [\%]} &  \multirow{2}{*}{mAP$_{70}$ [\%]} &  \multirow{2}{*}{mAP$_{75}$ [\%]}\\
state & & & \\
\hline
Before   &  99.4 & 99.4 & 97.2\\
After   & 99.3 & 99.3 & 97.0\\
\hline
\end{tabular}
\label{tab:detection}
\end{table}

\begin{table}[!t]
\setlength{\tabcolsep}{6pt}
\centering
\caption{Comparison of results for drivable area segmentation (MIoU -- Mean IoU, IoU -- Intersection over Union).}
\begin{tabular}{|c|c|c|c|} 
\hline
Quantisation  & \multirow{2}{*}{MIoU [\%]} & \multicolumn{2}{c|}{IoU [\%]}  \\
 \cline{3-4}
 state &  & Background & Drivable area \\ 
\hline
Before  & 97.31 & 97.88 & 96.75 \\
After  & 97.29 & 97.86 & 96.72\\
\hline
\end{tabular}
\label{tab:drivable_results}
\end{table}

\begin{table}[!t]
\setlength{\tabcolsep}{6pt}
\centering
\caption{Comparison of results for lane segmentation (MIoU -- Mean IoU, IoU -- Intersection over Union).}
\begin{tabular}{|c|c|c|c|} 
\hline
Quantisation & \multirow{2}{*}{MIoU [\%]} & \multicolumn{2}{c|}{IoU [\%]}  \\
\cline{3-4}
 state &  & Background & Lanes \\ 
\hline
Before  & 90.72 & 99.04 & 82.40 \\
After  & 90.69 & 99.04 & 82.33 \\
\hline
\end{tabular}
\label{tab:lane_results}
\end{table}

\begin{table}[!t]
\setlength{\tabcolsep}{6pt}
\centering
\caption{Comparison of results for object segmentation (MIoU -- Mean IoU, IoU -- Intersection over Union).}
\begin{tabular}{|c|c|c|c|c|c|c|c|} 
\hline
 \multirow{2}{*}{Quantisation} & \multirow{3}{*}{MIoU [\%]} & \multicolumn{6}{c|}{IoU [\%]}  \\
\cline{3-8}
 \multirow{2}{*}{state} &  & \multirow{2}{*}{Background} & \multirow{2}{*}{Pedestrian} & Amber & Red & Green & \multirow{2}{*}{Obstacle}\\ 
 &  &  &  & Light & Light & Light &  \\
\hline
Before  & 96.52 & 99.85 & 88.69 & 93.90 & 95.13 & 94.66 & 94.88 \\
After & 92.08 & 99.81 & 88.69 & 92.56 & 90.49 & 89.45 & 91.46 \\
\hline
\end{tabular}
\label{tab:seg_results}
\end{table}



\begin{table}[!t]
\setlength{\tabcolsep}{6pt}
\centering
\caption{Comparison of resource consumption on the Kria KV260 platform.}
\begin{tabular}{|c|c|c|c|c|c|c|c|}
\hline
\multirow{2}{*}{Resource usage} &  \multicolumn{4}{c|}{CPU cores} & \multirow{2}{*}{RAM} & \multirow{2}{*}{CMA} & \multirow{2}{*}{Power}\\
\cline{2-5}
{} & CPU$_0$ & CPU$_1$ & CPU$_2$ & CPU$_3$ & {} & {} & {}\\ 
\hline
Used & 85 \%  & 22 \%  & 3 \% & 3 \%  & 38 \%  & 6 \%  & 4.95 W\\
\hline
\end{tabular}
\label{tab:consumption}
\end{table}

\begin{table}[!t]
\setlength{\tabcolsep}{6pt}
\centering
\caption{Comparison of algorithm's performance on different computing platforms.}
\begin{tabular}{|c|c|c|c|c|c|c|}
\hline

\multirow{2}{*}{Embedded platform} & Power &  Speed & Execution & Model Inference \\
{} &  [W] & [fps] &  time [s] &  time [s] \\
\hline
Kria KV260 & 5 & 4.85 & 0.206 & 0.073\\
\hline
Nvidia Jetson Nano & 5  & 2.07 & 0.483 & 0.223\\
\hline
\multirow{2}{*}{Nvidia Jetson Xavier NX} & 10 & 4.35 & 0.230 & 0.093\\
{} & 20   &   5.48 & 0.182 & 0.068\\
\hline
\end{tabular}
\label{tab:platforms}
\end{table}


\section{Conclusion}
\label{sec:conslusion}

In this paper, we have discussed the implementation of a perception system for autonomous vehicles using a detection-segmentation network deployed in an SoC FPGA.
We have presented the process of preparing a custom dataset according to the requirements of the FPT'22 competition and the training of a neural network model.
We have also given a detailed description of the construction of a Mecanum wheel-based autonomous vehicle model, focusing on mechanical and electrical aspects.
A fully autonomous control algorithm has been implemented and run on the discussed platform, as well as on two eGPUs.
Several experiments have been performed, showing the efficiency and low power consumption of the proposed solution, which supports our thesis that the FPGA Kria KV260 using the MultiTask V3 neural network is a suitable solution for autonomous cars and robots with limited space and resources.


In future work, we will first refactor the code to further improve its efficiency. 
We also plan to test the vehicle model on the mock-up. 
Secondly, we will try to use the \textit{weakly supervised learning} and \textit{self-supervised learning} methods, which, in the case of an atypical, custom dataset, would allow a~significant reduction in the labelling process of the learning data.
\remove{We would also like to consider adding modules for depth estimation and optical flow, as these are often used in autonomous vehicle perception systems. }
\add{The next step would be to use the depth estimation branch in the MultiTask V3 model. This could be beneficial to further improve the vehicle's perception. Using the branch will require creating an additional dataset, based on the mock-up, as the pre-trained model did not give satisfactory results during testing. The dataset can be obtained by recording the environment using a depth sensor, such as the Intel RealSense L515, a device equipped with an RGB camera and a LiDAR sensor. We would also like to consider adding optical flow modules, as these are often used in autonomous vehicle perception systems.}

\section*{Acknowledgements}
\label{sec::acknowledgements}
The work presented in this paper was supported by the programme ``Excellence initiative - research university'' for the AGH University of Krakow.

\end{document}